\documentclass[preprint,12pt]{elsarticle}

\usepackage{amssymb}
\usepackage{amsthm}

\usepackage{amssymb}
\usepackage{color}
\usepackage{graphicx}
\usepackage{amsmath}
\usepackage{subfigure}
\usepackage{latexsym}

\usepackage{ulem}

\usepackage[ruled,vlined,linesnumbered]{algorithm2e}

\journal{Computer and System Sciences}

\begin{document}

\newtheorem{thm}{Theorem}
\newtheorem{prop}[thm]{Proposition}
\newtheorem{proposition}[thm]{Proposition}
\newtheorem{definition}[thm]{Definition}
\newtheorem{remark}[thm]{Remark}

\begin{frontmatter}



\title{Topological Tracking of Connected Components in Image Sequences}


\author{Rocio Gonzalez-Diaz$^1$,
Maria-Jose Jimenez$^1$,
Belen Medrano$^1$
}

\address{Applied Math (I) Dept., University of Seville, \\
Av. Reina Mercedes, s/n, 41012,  Seville, Spain\\
$\{$rogodi,majiro,belenmg$\}$@us.es
\\ \texttt{http://personal.us.es/$\{$rogodi,majiro,belenmg$\}$}
}

\begin{abstract}
Persistent homology provides information about the lifetime of homology classes along a filtration of cell complexes. Persistence barcode is a graphical representation of such information. A filtration might be determined by time in a set of spatiotemporal data, but classical methods for computing persistent homology do not respect the fact that we can not move backwards in time. In this paper, taking as input a time-varying sequence of two-dimensional (2D) binary digital images, we develop an algorithm for encoding, in the so-called {\it spatiotemporal barcode}, lifetime of connected components (of either the foreground or background) that are moving in the image sequence over time (this information may not coincide with the one provided by the persistence barcode).
This way, given a connected component at a specific time in the sequence, we can track the component backwards in time until the moment it was born, by what we call a {\it spatiotemporal path}. The main contribution of this paper with respect to our previous works lies in a new algorithm that
computes spatiotemporal paths directly,
valid for both foreground and background and developed in a general context, setting the ground for a future extension for tracking higher dimensional topological features in $nD$ binary digital image sequences. 
\end{abstract}

\begin{keyword}
Persistent homology\sep persistence barcodes\sep spatiotemporal data\sep binary digital image sequence analysis
\end{keyword}

\end{frontmatter}






\section{Introduction}

Persistent homology \cite{ELZ00,ZC05} and zigzag persistence \cite{Cd10} provide information about lifetime of homology classes along a filtration of cell complexes. Such a filtration might be determined by time in a set of spatiotemporal data. Our general aim is to compute the ``spatiotemporal" topological information of such filtration, taking into account that it is not possible to move backwards in time. This is not obvious if we use the known algorithms for computing (zigzag) persistent homology using time as filter function.

In the context of mobile sensor networks, \cite{dG06} is devoted to the problem of finding an {\it evasion} path that describes a moving intruder avoiding being detected by the sensors.  In \cite{dG06}, the region covered by sensors at time $t$ is encoded using a
Rips complex $R(t)$. A single cell complex $SR$ is computed by stacking the complexes $R(t)$ for all $t$. Th. 7 of \cite{dG06}
proves that there is no evasion path under a ``homological" criterion. Using zig-zag persistent homology, an equivalent condition is provided in \cite{adams15}.
A necessary and sufficient {\it positive cohomological criterion} for evasion in a general case is given in \cite{bib:GK17}.
Finally, in \cite{GCK15}, the authors analyze time-varying coverage properties in dynamic sensor networks by means of zigzag persistent
homology. 
 In all the mentioned papers, vertices represent sensors and edges are provided whenever two sensors can detect each other but their specific locations are unknown.

 We are concerned with the treatment of time-varying sequences of $nD$ binary digital images and the tracking of homology classes over time inspired by persistent homology methods. We deal with vertices at exact positions in each $nD$ image and adjacency relations between consecutive images are provided whenever there are cells in homologous positions
(that is, the  cells are in the same spatial position but at different times).  Our general goal is to compute a {\it spatiotemporal} barcode storing the evolution  of homology classes {\it over time}.
In this paper, we concentrate our effort in 2D images and track connected components of both the background and the foreground over time.
Roughly speaking, a spatiotemporal path will track a connected component over time and a spatiotemporal barcode will encode the evolution of homology classes over time. 
For example, the spatiotemporal barcode of the sequence of the four 2D binary digital images given in Fig.\ref{fig1-1}(a)-(d) reflects the fact that a connected component is born in the first image and dies in the third one, and another connected component is born in the second image and dies in the fourth one. 
This paper is an extension of our previous work \cite{iwcia2015}
in which we also focused on computing a spatiotemporal barcode for a time-varying sequence of 2D binary digital images. There, we used the construction of an algebraic-topological model (AT-model) \cite{GR05} to compute spatiotemporal paths.  In the present paper, though, we compute the spatiotemporal paths and barcode directly (without computing AT-models). 

We also  extend the definition of spatiotemporal filtration and spatiotemporal path to any dimension, what facilitates future extensions of our work to compute spatiotemporal $d$-barcodes in any dimension $d$.

Basics of persistent homology and AT-models are given in Section {\ref{sec:atmodels}. We introduce the problem of computing the
``correct" topological information of spatiotemporal data  through two simple examples in Section \ref{sec:problem}. Formal definitions to deal with a temporal sequence of cubical complexes are set in Section \ref{filtrations}. Our method to solve the problem is then introduced in Section \ref{sec:method}.
In Section \ref{nD}, we extend the definition of spatiotemporal path to any dimension. We conclude in Section \ref{sec:conclusions} and describe possible directions for future work.

\section{Persistent Homology through AT-models}\label{sec:atmodels}

Consider $\mathbb{Z}/2$ as the ground ring throughout the paper ( i.e. $1 + 1 = 0$). Roughly speaking, a cell complex $K$ is a general topological structure by which
a space is decomposed into basic elements (cells) of different dimensions that are
glued together by their boundaries (see the definition of CW-complex in \cite{bib:12}).
The dimension of a cell $\sigma\in K$ is denoted by $dim(\sigma)$.
   A cell $\mu\in K$ is a $d$-{\it face} of a cell $\sigma\in K$ if $\mu$ lies in the boundary of $\sigma$ and $d=dim(\mu)<dim(\sigma)$.
The cell complex $\partial K$ is built as follows: add a $d$-cell $\sigma$ of $K$ to $\partial K$ together with all its faces if $\sigma$ is face of exactly one $(d+1)$-cell in $K$.

If the cells in $K$ are $d$-dimensional {\it cubes} then $K$ is a {\it cubical complex}.
%
A $d$-dimensional cube ($d$-cube) is a product of $d$ elementary intervals $\prod_{i=1}^d I_i$.  An elementary interval is defined as a unit interval
$I=[k,k +1]$, with $k\in \mathbb{Z}$ or a degenerate interval $[k,k]$.
The number of non-degenerate
intervals in such product is the dimension of the cube. $0$-cubes, $1$-cubes, $2$-cubes and
$3$-cubes are vertices, edges, squares and 3D cubes (voxels) respectively. A cube $c_1$ is a face of a given cube $c_2$ if $c_1\subset c_2$. Given two
cubes, all the faces of a cube must also be a cube, as well as the intersection of any two cubes. A cubical complex has dimension
$D$ if the cubes are all of dimension at most $D$. The barycentric coordinates of a cube $c$ will be denoted by $r_c$.
}

 A
{\it $d$-chain} is a formal sum of $d$-cells in $K$. Since coefficients are either $0$ or $1$, we can think of a $d$-chain as a set of $d$-cells, namely those with coefficients equal to $1$. In set notation, the sum of two $d$-chains is their symmetric
difference.
The $d$-chains together with the addition operation form a group denoted as $C_d(K)$. Besides, the set $\{C_d(K)\}_{0\leq d\leq D}$,   is denoted by $C(K)$.
A set of homomorphisms $\{f_d: C_d(K)\to C_d(K')\}_{0\leq d\leq D}$, is called a {\it chain map} and denoted by $f:C(K) \to C(K')$. Given two $d$-cells $\sigma\in K$ and $\sigma'\in K'$, we say that $\sigma'\in f(\sigma)$ if
$\sigma'$ belongs to the $d$-chain $f_d(\sigma)$ (in set notation).
The {\it boundary map}
$ \partial: C(K) \to C(K)$ is defined on a $d$-cell $\sigma$ as the sum of its $(d-1)$-faces.
The $d$-chains with zero boundary form
a subspace $Z_d(K)$ of $C_d(K)$. The $d$-chains that are the boundary of $(d + 1)$-chains  form a subspace $B_d(K)$ of $Z_d(K)$. 
The
quotient group $H_d(K) = Z_d(K)/B_d(K)$ is the $d$-th homology group of $K$ (with $\mathbb{Z}/2$ coefficients).
The rank of $H_d(K)$, denoted by $\beta_d(K)$, is the $d$-th Betti number of $K$.
For a deeper introduction of these concepts, see \cite{Mun84,kaz,bib:12}.

A {\it filtration} of $K$ is an increasing sequence of cell complexes: $\emptyset =K_0\subset K_1\subset \cdots \subset K_{\ell}=K$.
The partial ordering given by such a filtration can be extended to a {\it total ordering} of the cells of $K$: $\{\sigma_1,\dots,\sigma_m\}$, satisfying that for each $i$, $1\leq i\leq m$, the faces of $\sigma_i$ lie on the set $\{\sigma_1,\dots,\sigma_i\}$.
The map $index: K\to \mathbb{Z}$ is defined by $index(\sigma_i):=i$.
Informally, the $d$-th persistent homology group \cite{ELZ00,ZC05} can be seen as a collection of $d$-homology classes (representing connected components when $d=0$, tunnels when $d=1$, cavities when $d=2$, ...)
that are born at or before we go from $K_{i-1}$ to $K_i$ and die after we go from $K_i$ to $K_{i+1}$. A persistence
$d$-{\it barcode}  \cite{bib:ghrist} is a graphical representation of the $d$-th persistent homology groups as a collection of horizontal line
segments ({\it bars}) in a plane.
Axis correspond to the indices of the cells in $K$.
For example, if a $d$-homology class is born at time $i$ (i.e. when $\sigma_i$ is added) and dies at time $j$ ($1\leq i<j\leq m$), then a bar $b=((i,i),(j,i))$
with endpoints
$(i,i)$ and $(j,i)$  is added to  the $d$-barcode.

In \cite{bib:caip2011} the authors establish a correspondence between the incremental algorithm for computing AT-models \cite{GR05} and the one for computing persistent homology
\cite{ELZ00}. 
More precisely, an {\it AT-model} for a cell complex $K$ is a quintuple $(f, g, \phi, K, H)$, where:

\begin{itemize}
\item $K$ is the cell complex.
\item $H$ is a set of cells of $K$ that describes the homology of $K$, in the sense that it contains a distinct $d$-cell for each $d$-homology class of a basis (see  next item), for all $d$.
The cells in $H$ are called {\it surviving cells}. For all $d$, the set of all the surviving $d$-cells
together with the addition operation form the group $C_d(H)$ which is  isomorphic to $H_d(K)$.
\item $g:  C(H) \rightarrow C(K)$
is a chain map  that maps each $d$-cell $h$ in $H$ to one representative cycle $g_d(h)$ of the corresponding homology class $[g_d(h)]$.
By a classical chain contraction property, it is true that $\{[g_d(h)]: h$ is a $d$-cell of $H\}$ is a basis for $H_d(K)$.
\item $f: C(K) \rightarrow C(H)$
is a chain map  that maps each $d$-cell  in $K$
to a sum of surviving cells, satisfying that  if $a,b\in C_d(K)$ are two homologous $d$-cycles then $f_d(a)=f_d(b)$.
\item $\phi: C(K) \rightarrow C(K)$
is a {\it chain homotopy} (see \cite{Mun84}).
 Intuitively, for a $d$-cell $\sigma$, $\phi_d(\sigma)$ returns a set of $(d+1)$-cells needed to be contracted to ``bring'' $\sigma$ to a surviving $d$-cell contained in $f_d(\sigma)$.
\end{itemize}

\section{Stating the Problem}\label{sec:problem}

Our general goal is to compute, for a time-varying sequence of $nD$ binary digital images, some kind of barcode that represents the evolution of homology classes {\it over time}.

Consider $\mathbb{Z}^n$ as the set of points with integer coordinates in $nD$ space $\mathbb{R}^n$. An {\it $nD$ digital binary image} is a set  $I=(\mathbb{Z}^n,\alpha,\beta,B)$, where $B\subset \mathbb{Z}^n$ is the {\itshape foreground}, $B^c=\mathbb{Z}^n \backslash B$ the {\itshape background}, and $(\alpha,\beta)$ is the adjacency relation for the foreground and background, respectively. In this paper we will deal with points with integer coordinates in 2D space $\mathbb{R}^2$, that is, {\it 2D digital binary images} (or 2D  images, for short),  $I=(\mathbb{Z}^2,8,4,B)$ (or $I=(\mathbb{Z}^2,B)$, for short), where $(8,4)$ is the adjacency relation for the foreground and background, respectively. All the 2D images considered here are finite, i.e., $I=({\cal D},B)$, where ${\cal D}\subset \mathbb{Z}^2$ is a finite domain, so $B\subseteq {\cal D}$ and $B^c={\cal D}\backslash B$ are both finite.

\begin{figure}[t!]
 \begin{center}
    \subfigure[ \label{fig1-1}]
    {\includegraphics[height=2cm]{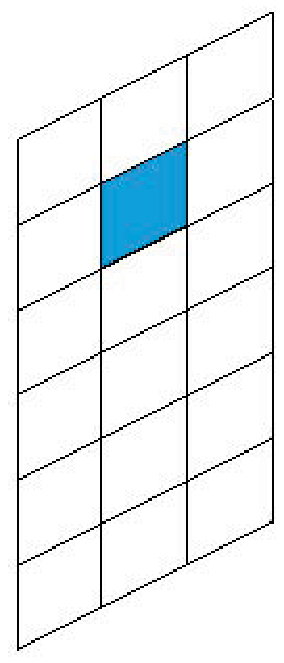}\;}
 \subfigure[ \label{fig1-2}]
  {\includegraphics[height=2cm]{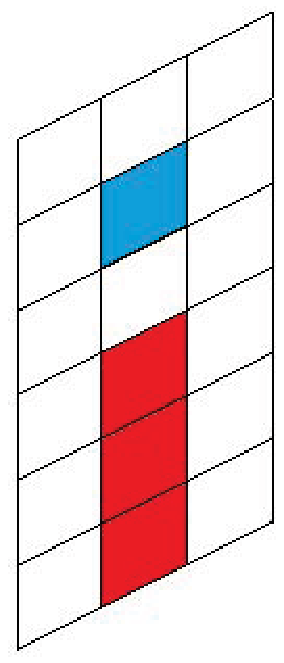}\;}
\subfigure[ \label{fig1-3}]
  {\includegraphics[height=2cm]{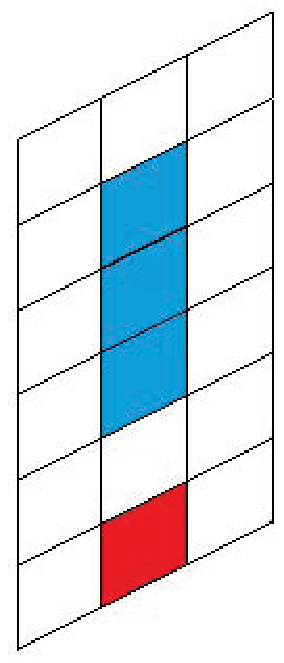}\;}
\subfigure[ \label{fig1-4}]
  {\includegraphics[height=2cm]{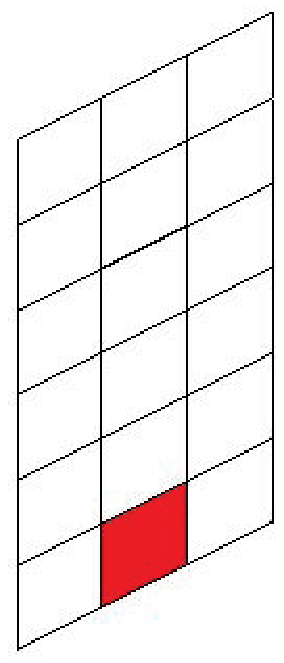}}
\qquad\qquad
\subfigure[ \label{fig2-1}]
   {\includegraphics[height=2cm]{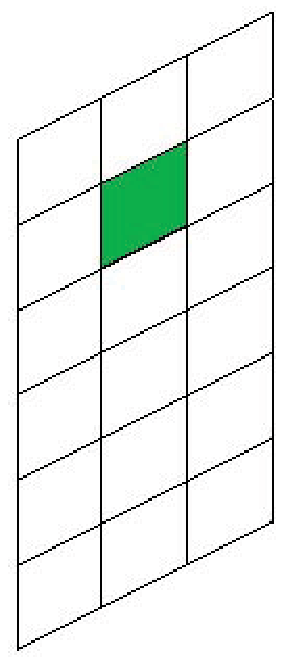}\;}
\subfigure[ \label{fig2-2}]
 {\includegraphics[height=2cm]{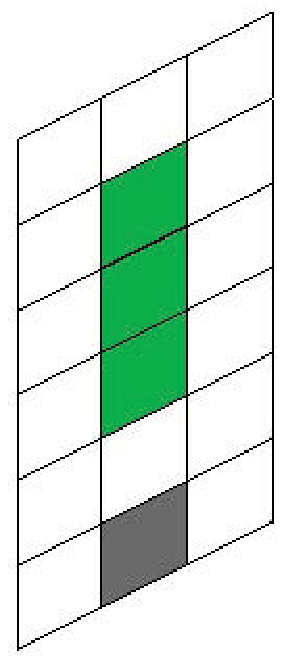}\;}
 \subfigure[ \label{fig2-3}]
  {\includegraphics[height=2cm]{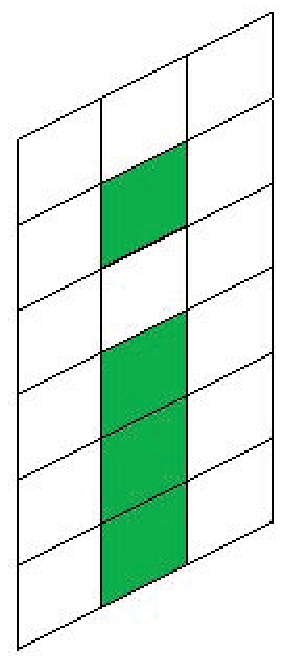}\;}
\subfigure[ \label{fig2-4}]
   {\includegraphics[height=2cm]{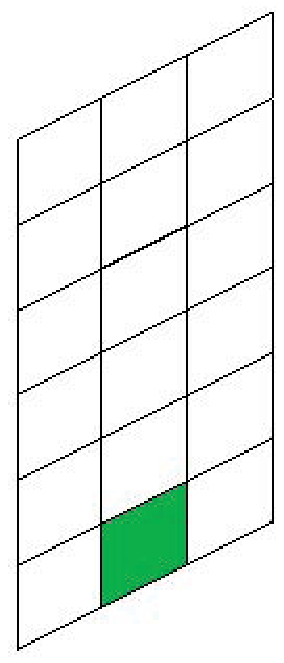}}
 \caption{Two different sequences of 2D images with pixels appearing and disappearing over time.  The reader is referred to the online version for color version of this figure. }
   \label{fig:p1}
  \end{center}
\end{figure}

   In order to give some intuition about the problem we want to state, let us consider the simple examples given in Fig. \ref{fig:p1}, in which two sequences of a few $4$-connected pixels appearing and disappearing over time, are shown.

\begin{figure}[t!]
\begin{center}
\subfigure[Complex $K'$. \label{fig3-1}]
{\includegraphics[width=6.5cm]{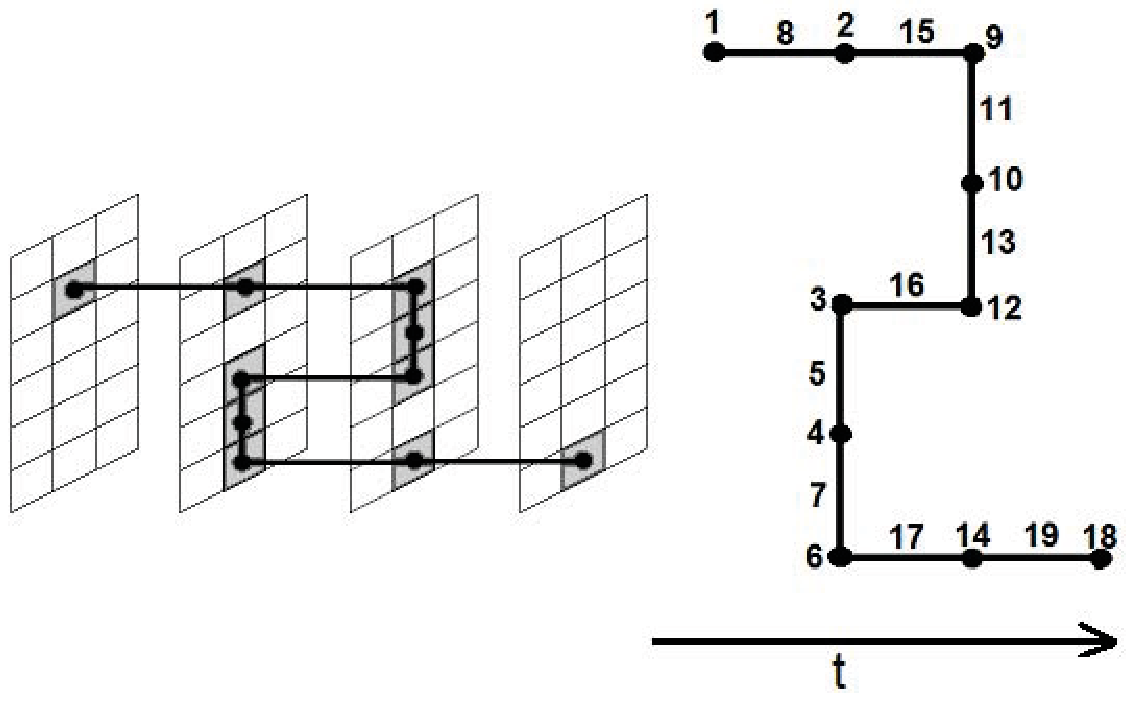}}
\subfigure[Complex $K''$. \label{fig3-2}]
{\includegraphics[width=6.5cm]{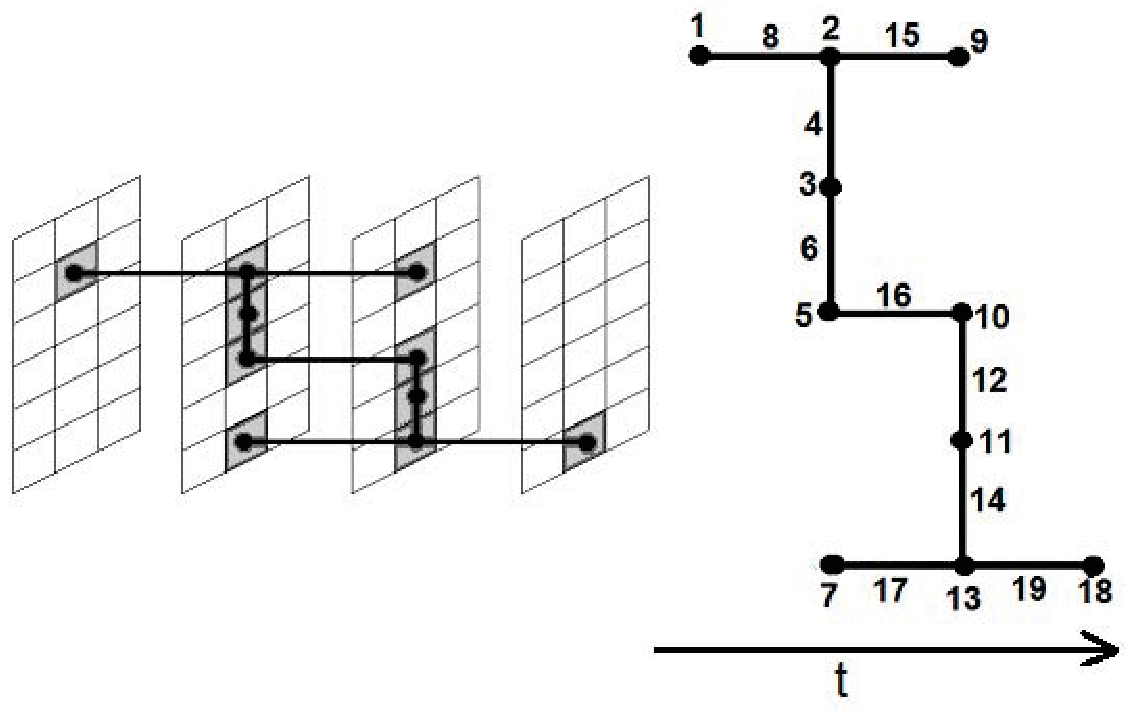}}
\end{center}
 \caption{Complexes $K'$ and $K''$ obtained, respectively, from the sequence shown in
Fig. \ref{fig:p1}(a)-(d) and Fig. \ref{fig:p1}(e)-(h).}
   \label{fig1}
\end{figure}

To encode the spatiotemporal information of the two sequences,
we construct two associated
  complexes (in fact, they are graphs) by
 replacing each pixel by a vertex and adding an edge between two vertices if:
\begin{itemize}
\item  The corresponding pixels are $4$-connected (in the same frame).
\item The vertices correspond to the same pixel (homologous coordinates) at consecutive  frames.
\end{itemize}
The resulting complexes $K'$ and $K''$ are shown in  Fig. \ref{fig1}.

Now, to compute persistent homology on these two complexes $K'$ and $K''$, we should select an appropriate filtration. Since we want to capture the variation of homology classes over time, we first classify the cells of $K'$ and $K''$ into {\it spatial} and {\it temporal} cells:
\begin{itemize}
\item  All vertices are spatial (since vertices represent pixels).
\item An edge is spatial if its endpoints (vertices) represent pixels of the same frame.
\item If an edge is not spatial then it is temporal.
\end{itemize}
Therefore, we have the following {\it spatial subcomplexes} of $K'$:
$G_1=\{1\}$, $G_2=\{2,3,4,5,6,7\}$, $G_3=\{9,10,11,12,13,14\}$, $G_4=\{18\}$.
 And the following sets of temporal cells:
$G_{1,2}=\{8\}$, $G_{2,3}=\{15,16,17\}$, $G_{3,4}=\{19\}$, where numbers correspond to the labels of cells shown in Fig \ref{fig1}(a).
The filtration $\emptyset=K'_0\subset K'_1\subset \cdots \subset K'_7=K'$ is obtained by interleaving the temporal cells after the correspondent  spatial subcomplexes. That is, $K'_1=G_1$, $K'_2=K'_1\cup G_2$, $K'_{3}=K'_2\cup G_{1,2}$, $K'_4=K'_3\cup G_3$, $K'_5=K'_4\cup G_{2,3}$, $K'_6=K'_5\cup G_4$ and $K'_7=K'_6\cup G_{3,4}$. 

The filtration of $K''$  is denoted with the same set of indices than the filtration of $K'$,  where numbers now correspond to the labels of the cells shown in Fig \ref{fig1}(b).

If we compute persistent homology of $K'$ and $K''$ using the above filtrations, we will obtain, in both cases, that a connected component ($0$-homology class) is born when cell  $1$ is added and survives until the end. So, in both cases, a bar with endpoints $(1,1)$ and $(19,1)$ is added to the persistence $0$-barcode.

However, we can observe that Fig. \ref{fig:p1}(a)-(d) cannot represent a connected component that is moving from the very beginning until the end while Fig. \ref{fig:p1}(e)-(h) can. So we wonder if we could modify the persistence $0$-barcode of the first sequence (Fig. \ref{fig:p1}(a)-(d)) so that it codifies the connected components that can survive along time. The idea is to replace the bar with endpoints $(1,1)$ and $(19,1)$ by respective bars from $(1,1)$ to $(13,1)$ and from $(3,3)$ to $(19,3)$, what will be formally described in next sections.

\section{Spatiotemporal filtrations and paths} \label{filtrations}

Going one step further, we  introduce here the concept of {\it spatiotemporal filtration} for a  more general setting of cubical complexes (in any dimension). The definition has been adapted from the construction described in \cite{dG06,adams15} for a sequence of simplicial complexes.

 First, given a sequence $S=\{Q_1,\dots,Q_{\ell}\}$ of cubical complexes, each one embedded in $\mathbb{R}^n$, we want to construct
 a  cubical complex embedded in $\mathbb{R}^{n+1}$,
 in which consecutive cubical complexes in $S$  are
stacked along a new (temporal) dimension.
\begin{definition}
Let $S=\{Q_1,\dots,Q_{\ell}\}$ be a sequence of cubical complexes embedded in $\mathbb{R}^n$. The  stacked cubical complex ${\cal SQ}[S]$ is the cubical complex embedded in $\mathbb{R}^{n+1}$ obtained as follows.
Initially,
${\cal SQ}[S]=\sqcup_{i=1}^{\ell} (Q_i\times \{i\})$. 
Now,
if a $d$-cell $\sigma$ with barycentric coordinates
$r_{\sigma}$
belongs to $Q_i\cap  Q_{i+1}$ (that is, $\sigma$ denotes corresponding cells in homologous positions in $Q_i$ and $Q_{i+1}$), for some $i$, $1\leq i<\ell-1$, add the $(d+1)$-cell
$\tau=\sigma\times [i,i+1]$  to ${\cal SQ}[S]$.
This way, the barycentric coordinates of $\tau$ are
$r_{\sigma}\times \{i+\frac{1}{2}\}$.
\end{definition}

\begin{figure}[t!]
\begin{center}
{\includegraphics[width=9cm]{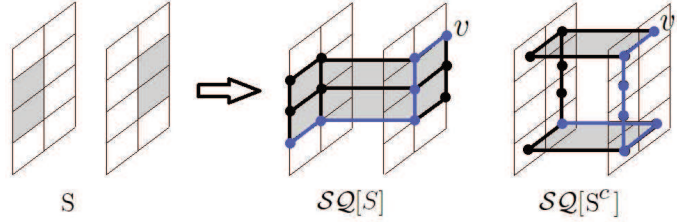}}
\end{center}
 \caption{A sequence $S$ of two 2D  images and the associated stacked cubical complexes $SQ[S]$ and $SQ[S^c]$ (the longest spatiotemporal paths obtained for $v$, in both cases, are drawn in blue). The reader is referred to the online version for color version of this figure.
 }
   \label{ejemplo2}
\end{figure}

See Fig. \ref{ejemplo2} as a toy example of stacked cubical complex. 
Since each cell $\sigma\in {\cal SQ}[S]$ can be identified by its  barycentric coordinates $r_{\sigma}\times \{ t_{\sigma}\}$,
then $\sigma$ is {\it spatial} if $t_{\sigma}\in\mathbb{Z}$; and it is {\it temporal} otherwise.

\begin{definition}
Let $ {\cal SQ}[S]$ be a stacked cubical complex  for a sequence of (spatial) cubical complexes $S=\{Q_1,\dots,Q_{\ell}\}$. Let  $Q_{i,i+1}$ denote the set of (temporal) cells with faces in both $Q_i$ and $Q_{i+1}$. The spatiotemporal  filtration $\emptyset = SQ_0 \subset SQ_1\subset \cdots \subset SQ_{m}={\cal SQ}[S]$
is given by: $SQ_1=Q_1$; $SQ_i=SQ_{i-1}\cup Q_{j+1}$ if $i=2j$ and $j>0$; and $SQ_i=SQ_{i-1}\cup Q_{j, j+1}$ if $i=2j+1$ and $j>0$.
\end{definition}

Once we have formally defined a spatiotemporal filtration for a sequence of cubical complexes in any dimension, we need to define {\it spatiotemporal} paths with the aim of setting down the restriction that it is not possible to move backwards in time.

First, recall that a {\it path} $p$ of edges in a cell complex $K$ from a vertex $v_0$ to a vertex $v_{q}$ is a chain of $1$-cells $p=e_1+e_2+\cdots + e_{q}$ (or $p=\{e_1,\dots,e_{q}\}$ in set notation) such that (1) only $e_1$ has $v_0$ as a face; (2) only $e_{q}$ has $v_{q}$ as a face; (3) each two edges $e_i$ and $e_{i+1}$ have a common face.

We have the following definition in the context of  stacked cubical complexes.

\begin{definition}\label{1}\cite{iwcia2015} Let $ {\cal SQ}[S]$ be a stacked cubical complex with spatiotemporal  filtration $\emptyset \subset SQ_0 \subset SQ_1\subset \cdots \subset SQ_{m}={\cal SQ}[S]$. A {\it spatiotemporal path} $p$ in ${\cal SQ}[S]$ is a path
such that  
the number of edges in $p\cap Q_{i,i+1}$ is less than or equal to $1$,
 for any $i$, $1\leq i <m$.
 
 Finally, two vertices are said to be  {\it spatiotemporally-connected} if there is a spatiotemporal path between them.
\end{definition}

Notice that, in a spatiotemporal path $p$, there are not two temporal edges in $p$ connecting the same consecutive spatial complexes of the sequence, which follows from the idea that it is not possible to move backwards in time.


\section{Topological Tracking of Connected Components in a 2D Image Sequence}\label{sec:method}

In this section, which is the main section of the paper, we explain how to compute spatiotemporal paths and barcodes for both the background and the foreground of a given  sequence of 2D  images.

We first need an adequate representation of 2D image sequences that captures the time-varying nature of the sequence.

Let $I=({\cal D},B)$ be a finite 2D image. We should deal with foreground and background differently,
since pixels in the foreground are $8$-connected while pixels in the background are $4$-connected. Regarding the foreground of $I$, a point  $p\in B$ can be interpreted as a unit closed square (called {\it pixel}) in $\mathbb{R}^2$ centered at $p$ with edges parallel to the coordinate axes.
The set of pixels centered at the points of $B$ together with their faces (edges and vertices) constitute a (2D) cubical complex denoted by $Q(I)$.
A cell $\sigma$ in $Q(I)$ can be identified
by its barycentric coordinates $r_{\sigma}\in \mathbb{R}^2$.

As for the background $B^c$ of $I$, the cubical complex
$Q(I^c)$ is computed as follows. Initially, $Q(I^c)=B^c$, which corresponds to the set of vertices of $Q(I^c)$.
Each two $4$-connected  vertices in $Q(I^c)$ form a unit edge that is added to $Q(I^c) $. Similarly, if four $4$-connected vertices in $Q(I^c)$ form a unit square $\sigma$, then $\sigma$ is also added to $Q(I^c)$.

\subsection{Computing Spatiotemporal Paths and Barcodes}
Consider now a sequence of 2D images $\{I_1,\dots,I_{\ell}\}$.
Let  S=$\{Q(I_1),$ $\dots,$ $Q(I_{\ell})\}$
and
$S^c=\{Q(I_1^c),\dots,Q(I_{\ell}^c)\}$
be the associated cubical complexes for, respectively, the foreground and the background of the 2D images in the sequence.
The stacked cubical complexes ${\cal SQ}[S]$ and ${\cal SQ}[S^c]$ (both embedded in $\mathbb{R}^3$) are computed as in Section \ref{filtrations}.  
See  Fig. \ref{ejemplo2} as an example of both ${\cal SQ}[S]$ and ${\cal SQ}[S^c]$.

In \cite{iwcia2015}, given a spatiotemporal filtration,
we designed an algorithm to compute the spatiotemporal barcode
encoding lifetime of connected components on the 2D image sequence over time.
The algorithm that we presented in that paper was based on the incremental algorithm for computing AT-models given in \cite{GR05}. 
In particular, the chain homotopy operator $\phi$ provides a path connecting each vertex $v$ to a distinguished vertex that represents the 
connected component that $v$ belongs to.
More specifically, in the algorithm given in \cite{iwcia2015}, 
a path $\phi(v)$ from $v$ to a surviving cell (vertex) was computed and, if $\phi(v)$ is not a spatiotemporal path, then it is broken into pieces that are spatiotemporal paths.
Regarding the spatiotemporal barcode, a bar was elongated at time $i$ if and only if $dim(\sigma_i)=1$ and the connected component that represents the bar is spatiotemporally connected to some of the  endpoints of the edge $\sigma_i$. Otherwise,  the bar was not elongated. This differentiates from classical persistence barcodes in which, for example, the bar corresponding to a connected component that appears at time $i$ and does not merge to other connected component later, is elongated until the very end.

In this paper, given a spatial connected component (i.e., a connected component in $I_j$ or $I^c_j$, $1\leq j\leq \ell$), we want to know in which previous frame it was born and
{\bf track
the connected component  evolution} along time.
For this aim, we compute a {spatiotemporal path $\phi'(v)$ for any vertex $v$ in
 ${\cal SQ}[S]$ or ${\cal SQ}[S^c]$ directly, i,e, without computing the chain homotopy $\phi$.
Notice that in order to pursue our goal in this paper, we only need the $1$-skeleton (i.e.  vertices and edges) of both
${\cal SQ}[S]$ and ${\cal SQ}[S^c]$.

\begin{algorithm}[h!]
  {\bf Input: } A spatiotemporal filtration.  associated to a sequence of cubical complexes. $\{Q_1,\dots,Q_{n}\}$ (either $Q_i=Q(I_i)$ for all $i$, or  $Q_i=Q(I^c_i)$ for all $i$, $1\leq i\leq n$).
 \\
%
Compute a total ordering $F=\{\sigma_1,\dots,\sigma_m\}$  of the vertices and edges of 
the given spatiotemporal filtration,
preserving the partial ordering given by the filtration.
\\
 Initialize sets $H'$, $TE$ and ${\cal B}$ as $\emptyset$; and maps $f'$ and $\phi'$ as zero.\\
\For{$i=1$ {\bf to} $m$}{
 \If{$\sigma_i$ is a vertex    }{
  $H':=H'\cup\{\sigma_i\}$ and $f'(\sigma_i):=\sigma_i$.\\
  Add the bar $((i,i),(i,i))$ to ${\cal B}$.
 }
 \If{$\sigma_i$ is an edge  and $f'\partial (\sigma_i)\neq 0$}{
   $TE:=TE\cup \{\sigma_i\}$.\\
   Let $\sigma_j\in  Q_s$ and $\sigma_{j'}\in Q_{s'}$ be the endpoints of $\sigma_i$ such that \\
   $\sigma_k=f'(\sigma_j)$ and $\sigma_{k'}=f'(\sigma_{j'})$ satisfy that $k'<k$.\\
   Let $r=\max\{s,s'\}$.  \\
   \If{$\sigma_k\in Q_r$ }{$H':=H'\setminus\{\sigma_k\}$.
   }
   {
   \For{$\ell=1$ {\bf to} $i-1$}{
   $\phi'_{aux}(\sigma_{\ell}):=\phi'(\sigma_{\ell})$;\\
     \If{$\sigma_{\ell}$ is a vertex in $Q_r$ and   $f'(\sigma_{\ell})=\sigma_k$}
     {     
     \If{$\phi'(\sigma_{\ell})+\phi'(\sigma_{j})
     +\sigma_i+\phi'(\sigma_{j'})$ is a spatiotemporal path
     from $\sigma_{\ell}$ to $\sigma_{k'}$ or $\sigma_{\ell}=\sigma_j$ } {$f'(\sigma_{\ell}):=\sigma_{k'}$,
$\phi'_{aux}(\sigma_{\ell}):=                   	\phi'(\sigma_{\ell})+\phi'(\sigma_j)+\sigma_i+\phi'(\sigma_{j'}).$
      }
     }
    }
    \For{$\ell=1$ {\bf to} $i-1$}{
   $\phi'(\sigma_{\ell}):=\phi'_{aux}(\sigma_{\ell}).$
   }
   }
  Add   the  bars $((k,k),(i,k))$ and $((k',k'),(i,k'))$ to ${\cal B}$.
 }	
}

{\bf Output: } The spatiotemporal paths $\phi'(v)$ for all the vertices $v$ in the filtration,  the associated spatiotemporal barcode ${\cal B}$ and
 the set of edges $TE$.
 \caption{Computing spatiotemporal paths and barcode of a spatiotemporal filtration.}
    \label{alg}
\end{algorithm}

Now, let us explain how Alg. \ref{alg}  works
\footnote{A naive implementation of the algorithm is available in http://grupo.us.es/cimagroup/}.
Let  $F$ be a total ordering of the cells of the spatiotemporal filtration considered.
We say that a cell $\sigma_i$ is older than a cell $\sigma_j$ if $i<j$.
Let us suppose that we are in step $i$ of the for-loop (line 4 in the algorithm). Then,  $H'$ is a collection of vertices of $F$ representing the spatiotemporally-connected components that were born and have survived until step $i$.
The map $f'$ connects each vertex $v\in F$ 
with the oldest vertex $w\in H'$ spatiotemporally connected to it. Besides, $\phi'(v)$ is a spatiotemporal path from $v$ to $w$.
At step $i$,  if $\sigma_i\in F$ is a vertex, then a new connected component is born (with only one vertex, $\sigma_i$), so $\sigma_i$ is added to $H'$, and $f'(\sigma_i)$ is updated.
If $\sigma_i$ is an edge and $f'(\partial\sigma_i)=0$, that means that
$f'(\sigma_j)=f'(\sigma_{j'})$ where $\sigma_j\in Q_s$ and $\sigma_{j'}\in Q_{s'}$ are the endpoints of $\sigma_i$.
Then, $\sigma_i$ is connecting two paths, $\phi'(\sigma_j)$ and $\phi'(\sigma_{j'})$, so no new connected component is created or destroyed. In fact, a new $1$-homology class  is born. See Fig. \ref{casos}.a.
Finally, if $\sigma_i$ is an edge and $f'(\partial\sigma_i)\neq 0$, then
 $\sigma_i$ is added to a set of edges $TE$ that we will be used later for tracking.
 Observe that $f'(\sigma_j)=\sigma_k$ and $f'(\sigma_{j'})=\sigma_{k'}$ for some $k,k'$ (see Prop. \ref{p}). We can suppose that $k'<k$.
 Let $r=\max\{s,s'\}$.
 If $s<s'=r$ then, we can not spatiotemporally connect $\sigma_j$ with $\sigma_{k'}$. See Fig. \ref{casos}.b.
If $s=s'=r$, then $\sigma_i$ is spatial and the  connected components represented by $\sigma_k$ and $\sigma_{k'}$ are spatiotemporally connected. If $s'<s=r$ and $\sigma_k\in Q_r$, then, again, the  connected components represented by $\sigma_k$ and $\sigma_{k'}$ are spatiotemporally connected.
Therefore we can remove $\sigma_k$ or $\sigma_{k'}$ from $H'$. We convene to remove the newest one which is  $\sigma_k$ (line 14 of the algorithm).
See Fig. \ref{casos}.c and \ref{casos}.e.
Finally, if $s'<s=r$ and $\sigma_k\not\in Q_r$, then we cannot remove $\sigma_k$. 
 Since $\phi'(\sigma_j)$ is a spatiotemporal path, then there only exists an edge $e$ in 
$Q_{s',s}\cap \phi'(\sigma_j)$.
See Fig. \ref{casos}.d.
Now, we update the spatiotemporal paths
of the vertices $\sigma_{\ell}$ spatiotemporally connected to $\sigma_k$  if and only if
$\sigma_{\ell}$ belongs to $Q_r$, to ensure that the updated path is also spatiotemporal (see lines 15-17 of the algorithm). See Fig. \ref{casos}.c and \ref{casos}.d.
Observe that we do not compute the AT-model $(F,H,f,g,\phi)$ for the spatiotemporal filtration  since
we are only interested in spatiotemporal paths and barcode.

Regarding time complexity of the algorithm, let $m$ be the number of cells of dimension $0$ and $1$ in the stacked cubical complex. Due to the  for-loops, the algorithm works in ${\cal O}(m^2)$ time. Regarding space, a spatiotemporal path $\phi'(v)$ must be stored for all the vertices $v$ in the filtration;  potentially, in the spatiotemporal barcode ${\cal B}$, there may be as many bars as vertices in the stacked complex; also potentially, any of  the edges of the stacked complex could be added to the set  $TE$, so  space complexity is ${\cal O}(m)$.

\begin{figure}[t!]
\begin{center}
\includegraphics[width=13cm]{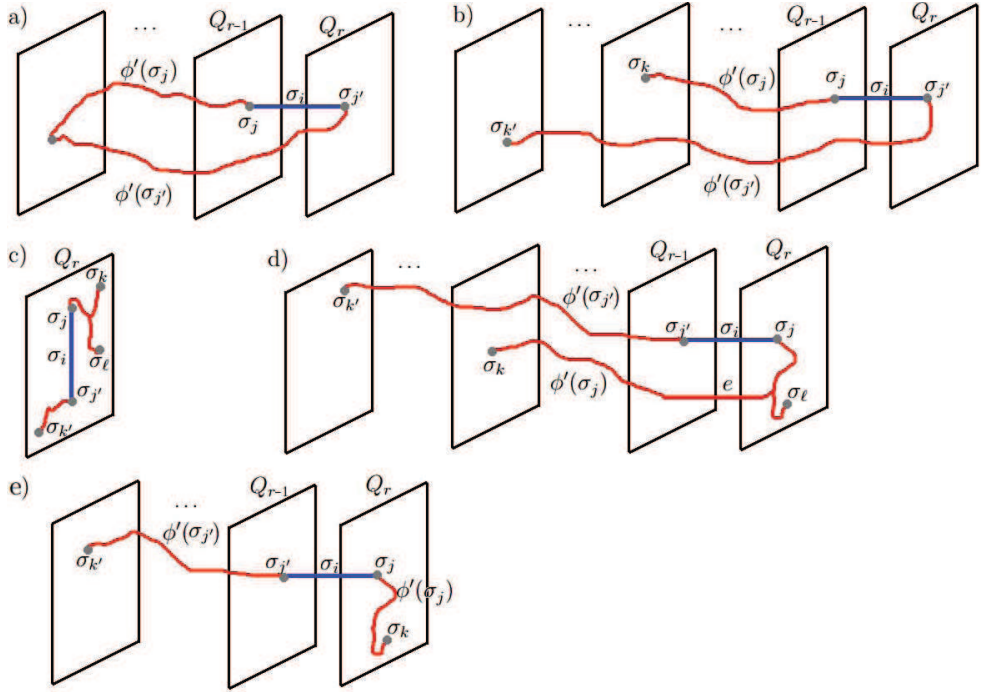}
\end{center}
 \caption{Different configurations when adding an edge $\sigma_i$: a) $f'(\partial\sigma_i)= 0$;    b) and d)
 $f'(\partial\sigma_i)\neq  0$ and  $\sigma_i$ is temporal;
 c) $f'(\partial\sigma_i)\neq  0$ and  $\sigma_i$ is spatial. e) $f'(\partial\sigma_i)\neq  0$, $\sigma_i$ is temporal and $\sigma_j,\sigma_k\in Q_r$. The reader is referred to the online version for color version of this figure.
 }
   \label{casos}
\end{figure}

\begin{prop}\label{p}
If $v$ is a vertex in $F$, then $f'(v)$ is a vertex $w$ in $F$.
\end{prop}

\proof
We will prove it by induction on the steps of Alg. \ref{alg}. Suppose we are in step $i$.
Then, by induction $f'(\sigma_{\ell})$ is a vertex if $\sigma_{\ell}$ is a vertex, for all $\ell$, $1\leq \ell<i$.
If $\sigma_i$ is a vertex then $f'(\sigma_i)$ is defined as $\sigma_i$, so the statement holds. If $\sigma_i$ is an edge, we eventually update $f'$ for some vertices $\sigma_{\ell}$, $\ell<i$, by $f'(\sigma_{\ell})=\sigma_{k'}$ which is a vertex, so the statement holds.
\qed

\begin{prop}
If $v$ is a vertex then $\phi'(v)$ is a spatiotemporal path.
\end{prop}

\proof
 We prove the statement by induction. Suppose that at step $i$ and for any vertex $\sigma_{\ell}\in F$, $1\leq \ell<i$,
  $\phi'(\sigma_{\ell})$ is a spatiotemporal path connecting $\sigma_{\ell}$ with another vertex $\sigma_k\in F$, being $1\leq k<\ell<i$.
  If $dim(\sigma_i)=0$, or $dim(\sigma_i)=1$ and $f'(\partial\sigma_i)=0$, then $\phi'(v)$ does not change for any vertex $v$ of $F$.
 If $dim(\sigma_i)=1$, $f'(\partial\sigma_i)\neq 0$ and $\sigma_j\in Q_r$ then, for a vertex $\sigma_{\ell}\in Q_r$ satisfying that either 
 $\phi'(\sigma_{\ell})+\phi'(\sigma_j)+\sigma_i+\phi'(\sigma_{j'})$ is a spatiotemporal path or $\sigma_{\ell}=\sigma_j$. So the statement holds.
\qed

Now, let us see how to ``topologically" track a connected component, once Alg. \ref{alg} has been executed. 
Let $v$ be a vertex in $Q_r$ for some $r$, $1\leq r\leq n$.
Recall that the spatiotemporal path $\phi'(v)$  connects $v$ with the oldest vertex $u$ in the sequence that is spatiotemporally connected with $v$.
In particular, $u$ represents the connected component $C$ in $Q_r$ that $v$ belongs to.
That is, we can detect the time in which the connected component $C$ is created  and we can follow it along the sequence. In fact, from the set of edges $TE$, we can obtain a directed tree $G_{TE}$ containing all the vertices
of ${\cal SQ}[S]$ 
that are spatiotemporally connected to other vertices in the sequence. 
Moreover, the directed paths in $G_{TE}$ ending at a vertex $v$ connect the vertex $v$ with {\bf all} the older vertices that are spatiotemporally connected to it.
See Fig. \ref{tree}.


 \begin{figure}[t!]
 \begin{center}
 \includegraphics[width=8cm]{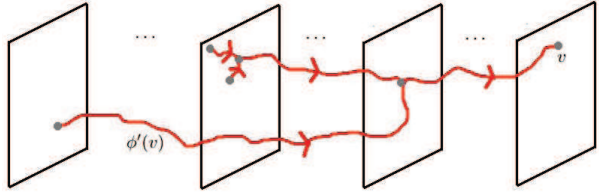}
 \end{center}  \caption{Spatiotemporal paths we can find in $G_{TE}$ for a given vertex $v$.}
    \label{tree}
 \end{figure}

\begin{figure}[t!]
\begin{center}
\subfigure
{\includegraphics[width=14cm]{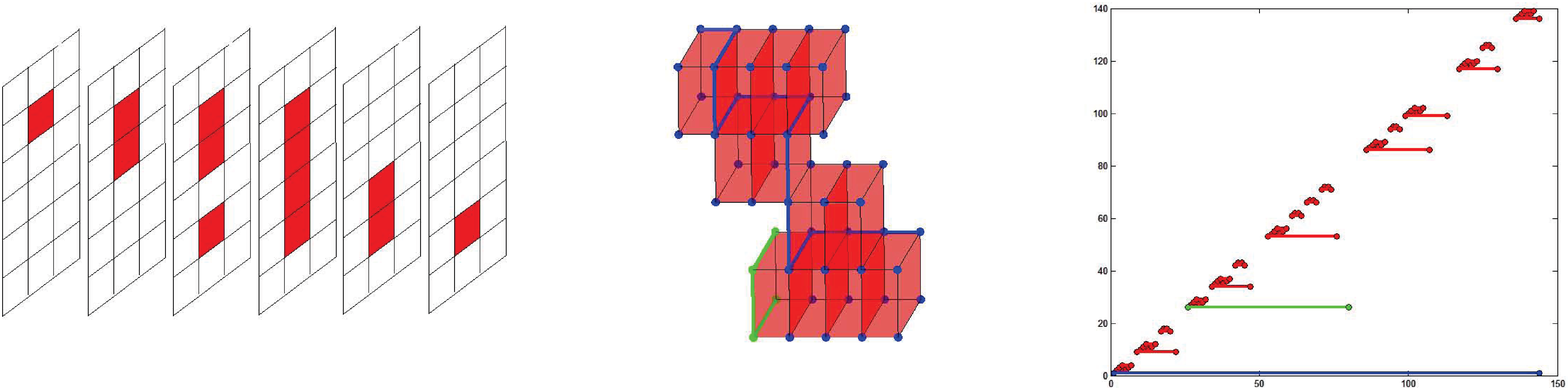}}
\subfigure
{\includegraphics[width=14cm]{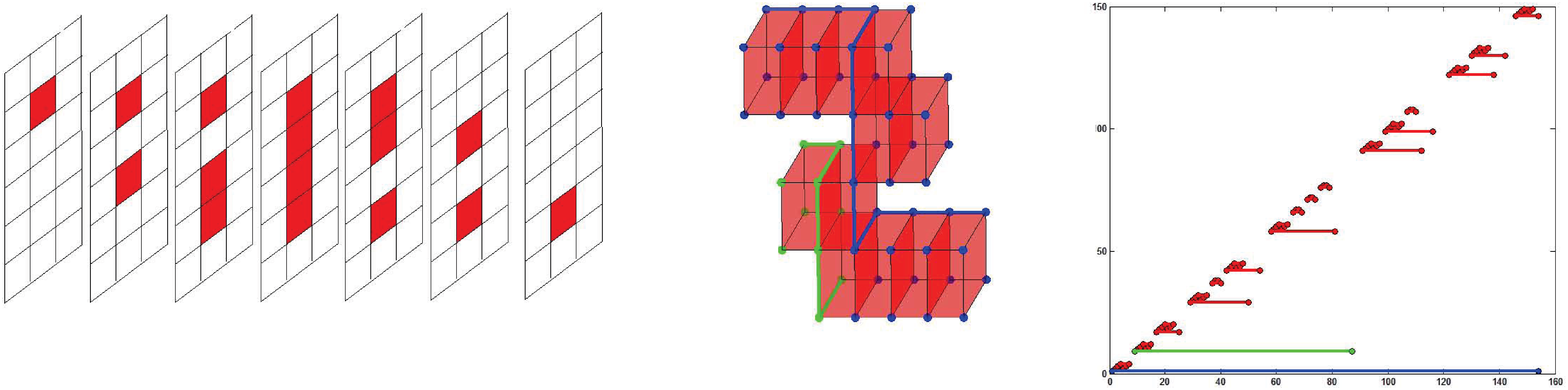}}
\subfigure
{\includegraphics[width=14cm]{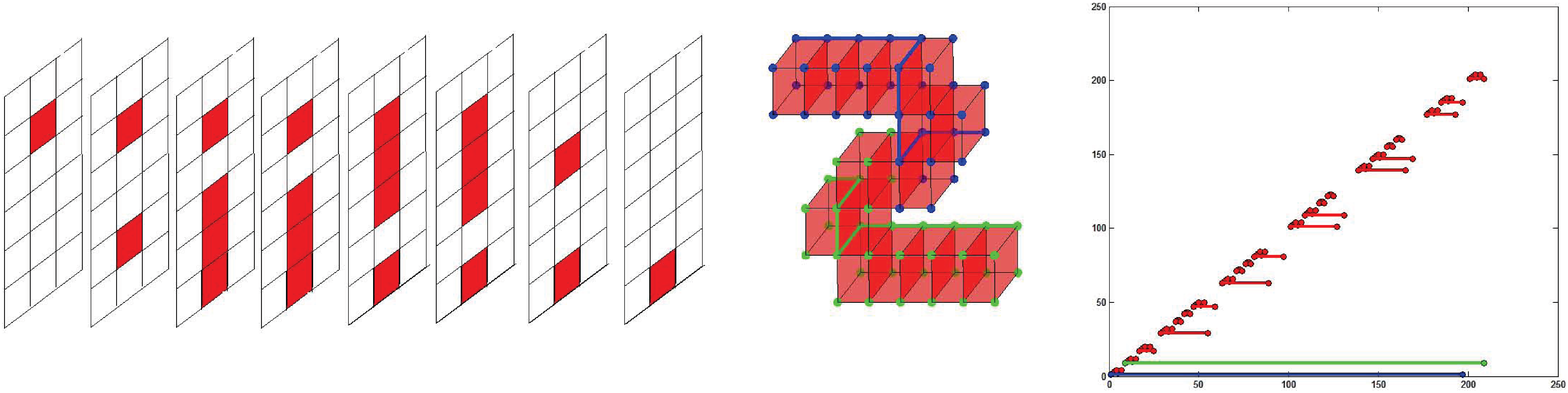}}
\end{center}
 \caption{Left: Three simple examples of 2D image sequences  ($t$ being the temporal dimension).
Center: Spatiotemporal paths of the longest-lived vertex in each spatiotemporally-connected component.   Rigth: The associated spatiotemporal barcodes. The reader is referred to the online version for color version of this figure.}
   \label{fig3}
\end{figure}

Fig. \ref{fig3} shows three simple examples of  2D image sequences.
The associated spatiotemporal barcodes are computed using Alg. \ref{alg}.
From left to right, the first and second spatiotemporal barcodes  have only one long bar,  while the third one has two. The longest spatiotemporal paths are pictured  in blue.
Notice that the classical persistence $0$-barcode would produce only one long bar in all of them.

\section{Towards the Computation of Spatiotemporal $d$-Barcodes for  $nD$ Image Sequences}\label{nD}

In this section, we generalize the concept of spatiotemporal path to higher dimension in order to set the ground for a future extension of Alg \ref{alg} for tracking higher dimensional topological features.

First, given a set of cells $T$ and a cubical complex $Q$, we denote by $T\cap Q$ the subcomplex of $Q$ obtained by taking all the cells of $T\cap Q$ together with all their faces.
Second, since the ground ring considered throughout the paper is $\mathbb{Z}/2$, we should restrict ourself to cubical complexes with torsion-free homology groups in order to extend the definitions of spatiotemporal paths and barcodes to $nD$.
Nevertheless, if the ground ring is $\mathbb{Z}$,  the definitions below are still valid for cubical complexes 
with non-torsion-free homology groups  replacing   $\mathbb{Z}/2$ by  $\mathbb{Z}$.

A ``homological'' $0$-path $P$ in a cubical
complex
$Q$ is defined as a subcomplex of $Q$ formed by a set of edges of $Q$ together with  their faces such that
\begin{itemize}
\item $\partial P=P_1\sqcup$\footnote{Recall that $A\sqcup B$ denotes the disjoint union of the sets $A$ and $B$.}$P_2$
for some subcomplexes $P_1$ and $P_2$ of $Q$,
\item $H_0(P)=H_0(P_1)=H_0(P_2)= \mathbb{Z}/2$,
 \item $H_j(P)=H_j(P_1)=H_j(P_2)=0$ for $j>0$.
 \end{itemize}

     \begin{definition}\label{2}
A {\it spatiotemporal $0$-path} $P$ in a spatiotemporal filtration
 for a given sequence of cubical complexes 
$S=\{Q_1,\dots,Q_{\ell}\}$ is a homological $0$-path in ${\cal SQ}[S]$ 
 such that
$H_0(P\cap Q_{i,i+1})=\mathbb{Z}/2$ 
and  $H_j(P\cap Q_{i,i+1})=0$  for $j>0$ and for all $i$, $1\leq i< \ell-1$.
 \end{definition}

 \begin{proposition}Def. \ref{1} and Def. \ref{2} are equivalent.
\end{proposition}

\proof
Def. \ref{2} $\Rightarrow$   Def. \ref{1}:
$P$ is formed by a set of edges and vertices. Since $H_0(P)=\mathbb{Z}$,
then $P$ has only one connected component. Since $H_j(P)=0$ for $j>0$, then $P$ does not contain any holes. 
So $P$ is a tree and 
$\partial P$ is a set of vertices.
Since
$H_0(P_1)=H_0(P_2)=\mathbb{Z}/2$ 
then both $P_1$ and $P_2$ can contain only one vertex. Therefore, $P$ is a path.
Besides, since $H_0(P\cap Q_{i,i+1})=\mathbb{Z}/2$, 
then the number of edges in $P\cap Q_{i,i+1}$ is less than or equal to $1$, for any $1<i \leq \ell-1$ (since edges in $ Q_{i,i+1}$ are disjoint).
\newline
Def. \ref{1} $\Rightarrow$  Def. \ref{2}:
Since $P$ is a path with no loops then $H_0(P)=\mathbb{Z}/2$, 
 $H_j(P)=0$ for $j>1$ and
$\partial P$ is formed by two vertices $v_0$ and $v_m$.
Then $P_1=\{v_0$\} and $P_2=\{v_m\}$. Besides, since the number of edges in $P\cap Q_{i,i+1}$ is less than or equal to $1$, for any $1<i \leq \ell-1$, then $H_0(P\cap Q_{i,i+1})=\mathbb{Z}/2$ 
and $H_j(P\cap Q_{i,i+1})=0$  for $j>0$ and for all $i$, $1\leq i< \ell-1$, what concludes the proof.
\qed

 The above definition has an easy generalization  to any dimension. A ``homological'' $d$-path $P$ in a cubical complex $Q$ is defined as a subcomplex of $Q$ formed by a set of $(d+1)$-cells of $Q$ together with  their faces such that
 \begin{itemize}
 \item $\partial P=P_1\sqcup P_2$
for some subcomplexes $P_1$ and $P_2$ of $Q$,
\item $H_i(P)=H_i(P_1)=H_i(P_2)= \mathbb{Z}/2$ 
for $i=0,d$,
 \item    $H_j(P)=H_j(P_1)=H_j(P_2)=0$  for $j\neq 0,d$.
    \end{itemize}
\begin{definition}A {\it spatiotemporal $d$-path} $P$ in a spatiotemporal filtration for a sequence of cubical complexes $S=\{Q_1,\dots,Q_{\ell}\}$ is a homological $d$-path in ${\cal SQ}[S]$
such that
$H_0(P\cap Q_{i,i+1})=H_d(P\cap Q_{i,i+1})=\mathbb{Z}/2$ 
and  $H_j(P\cap Q_{i,i+1})=0$ for $j\neq 0,d$ and for all $i$, $1\leq i\leq \ell-1$.
\end{definition}
  

\section{Conclusions and Future Work}\label{sec:conclusions}
In this paper, we have computed a modified persistence barcode, named spatiotemporal barcode, of a temporal sequence of 2D  images reflecting the time nature of the data. The computation can be made both for the foreground and background of the given images, what enables the tracking of $1$-holes of the foreground as (bounded) connected components of the background.
We have simplified the algorithm presented in \cite{iwcia2015} for computing spatiotemporal paths, avoiding the computation of  AT-models.
Although we have presented our algorithm for computing spatiotemporal paths only for 2D image sequences, it can be extended to sequences of images of any dimension, once a spatiotemporal filtration is constructed for the given sequence.
We have also extended the notion of spatiotemporal paths to any dimension.
This is part of an ongoing project to define and compute spatiotemporal $d$-barcodes (for any $d$) for sequences of $nD$  images.

\section*{Acknowledgments}
This research has been partially supported  by MINECO, FEDER/UE under grant MTM2015-67072-P.

We would like to thank the reviewers for valuable suggestions and comments.
%
%


\end{document}